\documentclass[runningheads]{llncs}
\usepackage{amsmath} 
\usepackage[T1]{fontenc} 
\usepackage{graphicx}
\usepackage{amsfonts} 
\usepackage{xcolor}
\begin{document}
\title{Normative Diffusion Autoencoders: Application to Amyotrophic Lateral Sclerosis}
\titlerunning{Normative Diffusion Autoencoders}
\authorrunning{Ijishakin et al.}
\author{Ayodeji Ijishakin\inst{1} \and Adamos Hadjasavilou\inst{1} \and Ahmed Abdulaal\inst{1} \and Nina Montana-Brown\inst{1} \and Florence Townend\inst{1} \and Edoardo Spinelli\inst{3,4,5} \and Massimo Filippi\inst{3-7} \and Federica Agosta\inst{3,4,5} \and James Cole\inst{1} \and Andrea Malaspina\inst{2}}
\institute{Centre for Medical Image Computing, Department of Computer Science, University College London, UK \and Department of Neuromuscular Diseases, University College London, UK\and  Neuroimaging Research Unit, Division of Neuroscience, IRCCS San Raffaele Scientific Institute, Milan, Italy \and Vita-Salute San Raffaele University, Milan, Italy, \and Neurology Unit, IRCCS San Raffaele Scientific Institute, Milan, Italy; \and Neurorehabilitation Unit, IRCCS San Raffaele Scientific Institute, Milan, Italy \and Neurophysiology Service, IRCCS San Raffaele Scientific Institute, Milan, Italy.
 \\\email{Corresponding Email: ayodeji.ijishakin.21@ucl.ac.uk}} 
\maketitle              % typeset the header of the contribution
\begin{abstract}
Predicting survival in Amyotrophic Lateral Sclerosis (ALS) is a challenging task. Magnetic resonance imaging (MRI) data provide in vivo insight into brain health, but the low prevalence of the condition and resultant data scarcity limit training set sizes for prediction models. Survival models are further hindered by the subtle and often highly localised profile of ALS-related neurodegeneration. Normative models present a solution as they increase statistical power by leveraging large healthy cohorts. Separately, diffusion models excel in capturing the semantics embedded within images including subtle signs of accelerated brain ageing, which may help predict survival in ALS. Here, we combine the benefits of generative and normative modelling by introducing the normative diffusion autoencoder framework. To our knowledge, this is the first use of normative modelling within a diffusion autoencoder, as well as the first application of normative modelling to ALS. Our approach outperforms generative and non-generative normative modelling benchmarks in ALS prognostication, demonstrating enhanced predictive accuracy in the context of ALS survival prediction and normative modelling in general. 
\end{abstract}
\section{Introduction}
ALS is a rare and incurable neurodegenerative disease, with a global prevalence of 4.36 per 100,000 people \cite{globalprev,Knibb2016AALS}. Access to ALS patient data is limited which makes predicting survival time particularly challenging \cite{ijishakin2023semi}. This is compounded by the high mortality rate associated with the disease (median survival time of 3 years), which further limits the length of the associated natural history \cite{Swinnen2014TheSclerosis}. Additionally, ALS is characterised by subtle neuroanatomical changes which are hard to detect within neuroimaging data \cite{agosta2010present}. As such, to produce effective predictive models of ALS survival, we need approaches that increase statistical power and are sensitive to the neuroanatomical changes associated with ALS disease.  
\begin{figure*}[t]
\centering
    \includegraphics[scale=0.1]{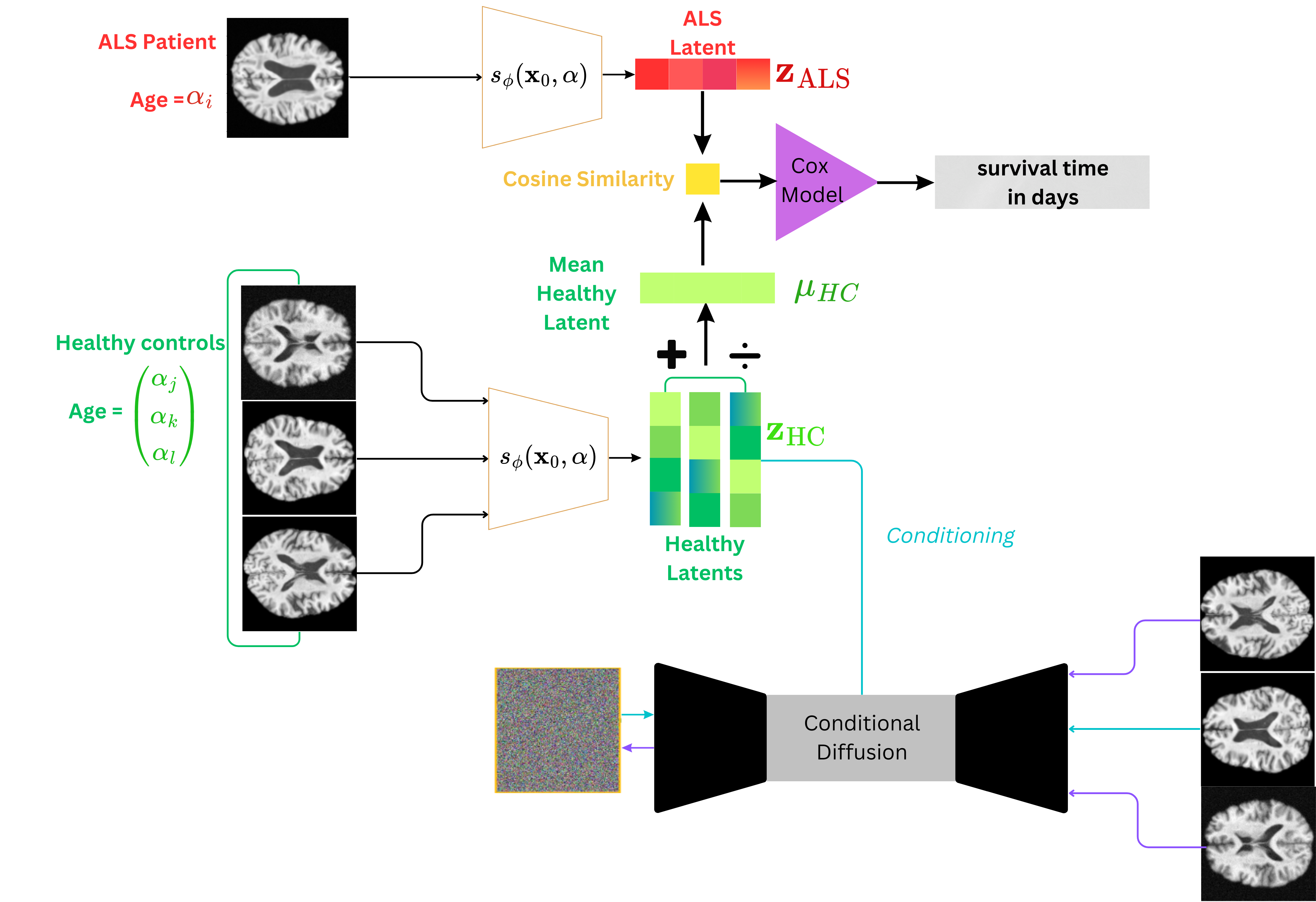}
\caption{The normative diffusion autoencoder. We first train our hierarchical diffusion autoencoder, $p_{\theta}(\mathbf{x}_{0:T}| \mathbf{z})$ on a healthy cohort. Then we use $ s_{\phi}(\mathbf{x}_{0}, \alpha)$ to compute the latent representations, of all MRIs from our healthy cohort (N=4120), conditional on their ages $\alpha$. Next, we average these representations to produce $\boldsymbol{\mu}_{HC}$. After which we compute the latent representation of MRIs from ALS patients to produce $\mathbf{z}_{ALS}$. We quantify the similarity to the healthy cohort by computing the cosine similarity between $\boldsymbol{\mu}_{HC}$ and $\mathbf{z}_{ALS}$. Finally, we fit a Cox model using this similarity metric to model its effect on survival time in days.} 
\label{Braggie}
\end{figure*} 
One such solution is normative modelling (NM), which estimates the probability distribution of a healthy cohort, such that deviations from this distribution may be quantified and used to make inferences about clinical outcomes \cite{bozek2023normative,wolfers2018mapping,zabihi2019dissecting}. As these models pool large healthy reference cohorts they have increased statistical power and therefore sensitivity to atypical data points \cite{rutherford2022charting,bethlehem2022brain}. Whilst NM for neuroimaging has been used across neurology and psychiatry, its utility in ALS has not been investigated \cite{borghi2006construction}. Further, most normative modelling approaches are restricted to capturing linear relationships within scalar/low-dimensional datasets \cite{bozek2023normative,marquand2019conceptualizing,yee2015vector}. On the other hand, deep generative models capture non-linear and semantically rich relationships in high-dimensional data. In particular, deep denoising diffusion models (diffusion models) are the state-of-the-art generative model for their log likelihoods and fretchet-inception distances (image synthesis quality) \cite{Ho,song}. Diffusion autoencoders extend their representation learning capacity making them a natural framework to pair with NM in the context of modelling a complex disease like ALS with MR data \cite{DiffAE}. 

In this work, we present the first application of NM with a diffusion autoencoder. To sensitively model healthy neuroanatomy, we introduce a new class of diffusion model, called the hierarchical diffusion autoencoder, and demonstrate its ability to produce rich latent space representations of normative brain structure. We illustrate the value of our model over both generative and non-generative NM techniques in producing a metric that is useful for survival prediction in ALS disease.
\section{Related Work} 
Wang et al. (2023) \cite{wang2023} and Lawry et al. (2022) \cite{ana2022} both used conditional Variational Autoencoders (VAEs) to produce normative models which were applied to Alzheimer's disease classification. Other work has also used multi-modal VAEs to conduct normative modelling of Alzheimer's Disease across multiple MR modalities \cite{ana2023}. However, VAEs constrain the latent normative distribution by placing a Gaussian prior on it and they are no longer the state-of-the-art generative model. Pinaya et al. (2022) \cite{Walter}, Behrendt et al. (2023) \cite{patchy-diffusion}, and Iqbal et al. (2023) \cite{iqbal} used latent diffusion, patch-based diffusion, and masked diffusion models respectively to conduct anomaly detection on CT and MR scans from individuals with multiple sclerosis lesions and brain tumours. Although these approaches are normative, in that they detect abnormalities based on a healthy distribution, these models are designed to aid tumour and lesion segmentation, so they do not naturally extend to survival analysis in ALS. Additionally, these approaches produce image-level normative deviations, which may not be as effective as latent space deviations, for a disease like ALS with very subtle neuroanatomical changes.
\section{Background}
Diffusion autoencoders learn a data distribution, $q(\mathbf{x}_{0})$, via two steps, (1) the forward process and (2) the reverse process. The forward process,  $q(\mathbf{x}_{1:T}|\mathbf{x}_{0})$, corrupts our data $\mathbf{x}_{0} \sim q(\mathbf{x}_{0})$ by adding Gaussian noise in a Markovian fashion with the following form: 
\begin{equation}
    q(\mathbf{x}_{t}|\mathbf{x}_{t-1}) := \mathcal{N}(\mathbf{x}_{t};\sqrt{1-\beta_{t}} \mathbf{x}_{t-1}, \beta_{t} \textbf{I}), \quad     q(\mathbf{x}_{1:T}|\mathbf{x}_{0}) := \prod^{T}_{t=1} q(\mathbf{x}_{t}|\mathbf{x}_{t-1}) 
\end{equation} 
Where $T$ is the total number of noising steps and $\{\beta_{t}\}_{t=1}^{T}$ are scalars that control the variance over the forward process. Following $T$ noising steps our data should approach a standard normal, $\mathbf{x}_{T} \sim \mathcal{N}(\textbf{0}, \textbf{I})$. 
% We may also obtain the noised version of our image at time $t$ via: 
% \begin{equation}
% q(\mathbf{x}_{t}|\mathbf{x}_{0}) = \mathcal{N}(\sqrt{\alpha_{t}}\mathbf{x}_{0},(1 - \alpha_{t})\mathbf{I}), \quad  \text{where} \quad  \alpha_{t} = \prod_{s=1}^{t}(1 - \beta_{s}) 
% \label{Eqn2}
% \end{equation} 
The reverse process, $p_{\theta}(\mathbf{x}_{0:T} | \mathbf{z})$, recovers our data distribution conditional on a deterministic semantic latent $\mathbf{z}$. The reverse process is similarly a Markov chain that starts at the maximum level of data corruption $p(\mathbf{x}_{T}) = \mathcal{N}(\mathbf{x}_{T}; \textbf{0}, \textbf{I})$ and ends at our model distribution $p_{\theta}(\mathbf{x}_{0}|\mathbf{z})$.
\begin{equation} 
    p_{\theta}(\mathbf{x}_{0:T}|\mathbf{z}) := p(\mathbf{x}_{T})\prod_{t=1}^{T}p_{\theta}^{(t)}(\mathbf{x}_{t-1}| \mathbf{x}_{t}, \mathbf{z}), \quad     p_{\theta}(\mathbf{x}_{0} | \mathbf{z}) = \int p_{\theta}(\mathbf{x}_{0:T} | \mathbf{z})d\mathbf{x}_{1:T}
\end{equation}
We obtain $p_{\theta}(\mathbf{x}_{0}|\mathbf{z})$ via a latent variable model with two types of latents. Our stochastic latents $\{\mathbf{x}_{t}\}_{t=1}^{T} \in \mathbb{R}^{D}$  which model the low-level random aspects of our data. And our semantic latent $\mathbf{z} \in \mathbb{R}^{d}$ (typically output by a neural network $s_{\phi}(\mathbf{x}_{0})$), which models the high-level semantics of our data. The stochastic latent variables share the size of our data, $\mathbf{x}_{0} \in \mathbb{R}^{D}$ whilst $d << D$. 

After $t$ noising steps, (and due to reparameterisation), $\mathbf{x}_{t} = \sqrt{\alpha_{t}}\mathbf{x}_{0} + \sqrt{1 - \alpha_{t}}\epsilon$ where $\epsilon \sim \mathcal{N}(\mathbf{0}, \mathbf{I})$ \cite{Ho} and $\alpha_{t} = \prod_{s=1}^{t}(1 - \beta_{s}$). As such, we may learn a neural network $\epsilon_{\theta}^{(t)}(\mathbf{x}_{t}, \mathbf{z})$ which predicts the noise, $\epsilon$ that was sampled when noising the image. Our training objective is an MSE:  $||\epsilon^{(t)}_{\theta}(\mathbf{x}_{t}, \mathbf{z}) - \epsilon||^{2}_{2}$. 

\subsection{Cox Proportional Hazard Models}
Cox proportional hazard models of survival assess the impact of $n$ covariates,  $x_{1},...,x_{n}$, on the likelihood of death occurring at a time $t$. This is calculated with the hazard function $h(t)$: 
\begin{equation}
h(t) = h_{0}(t)e^{b_{1}x_{1} + b_{2}x_{2} ... + b_{n}x_{n}}    
\end{equation}
Where $b_{1},...,b_{n}$ are the $n$ coefficients which scale the covariates relative to the amount of influence they have on survival, $ \{ e^{b_{i}} \}_{i=1}^{n}$ are the  \textit{hazard ratios} and $h_{0}$ is the \textit{baseline hazard} when all $x_{i} = 0$. If $b_{i}$ > 0 (and thus $e^{b_{i}}$ > 1) it suggests that as the corresponding covariate, $x_{i}$, increases the hazard rate also increases. When $e^{b_{i}}$ < 1, $x_{i}$ is negatively associated with risk of mortality.
\section{Method}
\subsection{Hierarchical Diffusion Autoencoders}
\begin{figure}[]
\centering
    \includegraphics[scale=0.6]{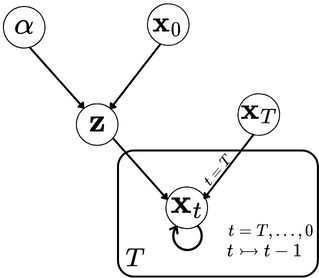}
\caption{The graphical model induced by our hierarchical diffusion autoencoder. The conditional variable $\alpha$ condition's the semantic latent $\mathbf{z}$ alongside the original image $\mathbf{x}_{0}$. The reverse process then iteratively denoises starting from our final stochastic latent $\mathbf{x}_{T}$ conditional on $\mathbf{z}$ to recover $\mathbf{x}_{0}$.} 
\label{GraphicalModel}
\end{figure}

A strong association between ageing and increased risk of neurodegenerative diseases has been established, and many studies suggest that the older the age of ALS patients at the onset of symptoms, the shorter the survival time \cite{globalprev,Knibb2016AALS,ijishakin2023semi}. To ensure that the representations produced by our model are well-conditioned on how healthy neuroanatomy changes over the lifespan, we developed a new class of diffusion model called the hierarchical diffusion autoencoder. The semantic encoder of the model, $s_{\phi}(\mathbf{x}_{0}, \alpha)$, also takes as input a conditional variable, $\alpha$. In the present case, $\alpha$ is the age associated with the MR image but it could be any conditioning variable. By conditioning the semantic latent, $\mathbf{z}$, on age, it induces the graphical model seen in Figure \ref{GraphicalModel}. Given that $\alpha$ conditions $\mathbf{z}$, which conditions the reverse process $p_{\theta}(\mathbf{x}_{0:T}|\mathbf{z})$, the model uses a hierarchy of conditioning, hence the name hierarchical diffusion autoencoder.  
\subsection{Latent Similarity Metric and Survival Analysis}
Once our hierarchical diffusion autoencoder model is trained, we create a vector, $\boldsymbol{\mu}_{HC}$, which is the average representation of the healthy cohort within our semantic latent space:
\begin{equation}
    \boldsymbol{\mu}_{HC} = \frac{1}{N}\sum_{i=1}^{N}\mathbf{z}_{HC,i}
\end{equation} 
Where $N$ is the size of our healthy cohort, and $\{ \mathbf{z}_{HC} \}_{i=1}^{N}$ are the latent representations corresponding to each individual in our healthy cohort (hence the subscript HC). We compute the similarity between the latent representation of an MR image from the $i$th ALS patient, $\mathbf{z}_{ALS, i}$ as the cosine similarity between $\boldsymbol{\mu}_{HC}$ and $\mathbf{z}_{ALS, i}$. 
\begin{equation}
    \operatorname{sim}\left( \boldsymbol{\mu}_{HC}, \mathbf{z}_{ALS, i} \right)= \frac{\boldsymbol{\mu}_{HC} \mathbf{z}_{ALS, i}^{{\mathsf{T}} }}{\| \boldsymbol{\mu}_{HC} \|  \| \mathbf{z}_{ALS, i} \|}
\end{equation} 
Previous latent normative modelling works have used the Mahalanobis distance as the deviation/similarity metric \cite{ana2022,ana2023}. We instead opt for computing the cosine similarity to $\boldsymbol{\mu}_{HC}$ due to its simplicity and computational efficiency. 
% The Mahalanobis distance requires the calculation of the precision matrix for the distribution of latent representations of the whole healthy cohort. This is computationally intensive when the training set size grows, and is prone to numerical overflow errors because it involves a matrix inversion. As such, 
Our approach is a simple and lightweight alternative, with an equally intuitive interpretation. Namely, the more similar a given $\mathbf{z}_{ALS}$ is to $\boldsymbol{\mu}_{HC}$, the better survival outcomes the corresponding patient should have. After computing our latent similarity metric we fit Cox models to analyse its influence on survival time, defined as the length of days between the acquisition of an MR scan and the individual's death. Figure \ref{Braggie} shows the full model pipeline.
\begin{figure}[t]
    \centering 
    \includegraphics[scale=0.06]{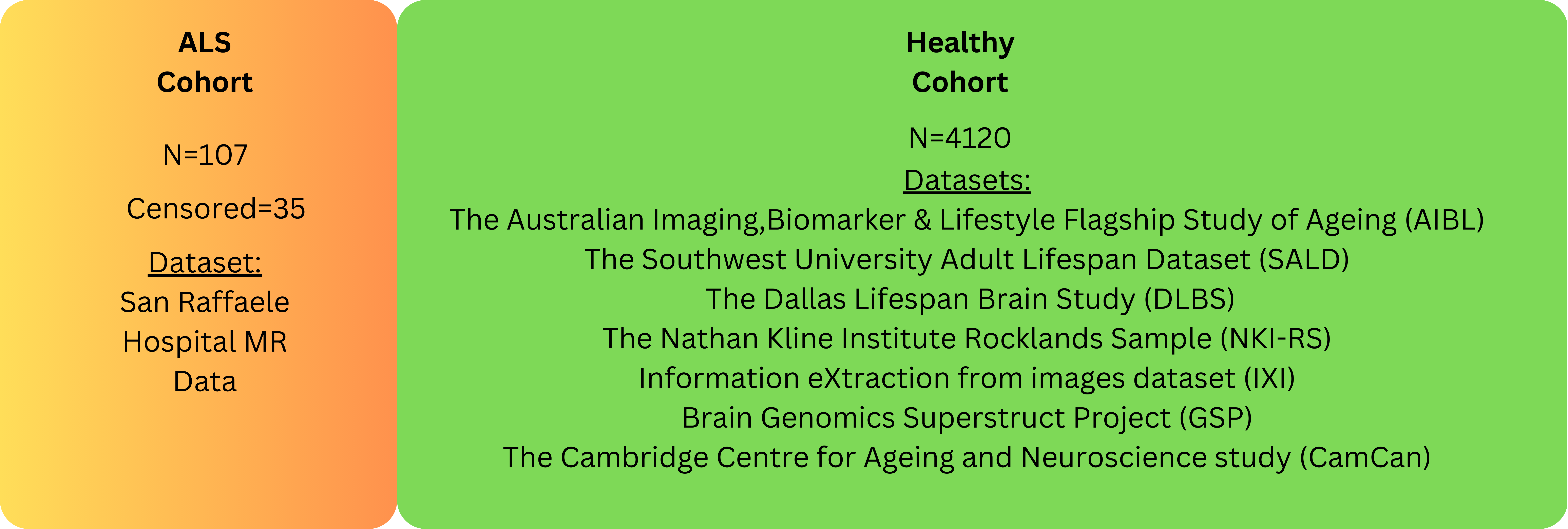}
    \caption{Our dataset. On the left is a description of our ALS cohort, and on the right are the names of the 7 datasets that comprised our healthy cohort.}
    \label{dataset}
\end{figure} 
\section{Experiments} 
\subsection{Datasets}
Figure \ref{dataset} shows the full breakdown of our datasets. Our healthy cohort was n=4120 participants with 2D structural T1-weighted MRIs (mean age = 41.7 years, std = 21.7 years) derived from 7 public datasets. These were: The Australian Imaging Biomarker \& Lifestyle Flagship Study of Ageing (AIBL); The Southwest University Adult Lifespan Dataset (SALD); The Dallas Lifespan Brain Study (DLBS), The Nathan Kline Institute Rocklands Sample (NKI-RS); Information eXtraction from images dataset (IXI); Brain Genomics Superstruct Project (GSP) and the Cambridge Centre for Ageing and Neuroscience Study (CamCan). Our ALS cohort was composed of n=107 patients with 2D structural T1-weighted MRIs (mean age = 61 years, std = 10.9 years) from the San Raffaele Hospital in Milan. There were 35 censored patients (72 deceased) and the maximum follow-up period was 9.2 years. 
\subsection{Pre-processing}
\begin{figure}
    \centering 
    \includegraphics[scale=0.06]{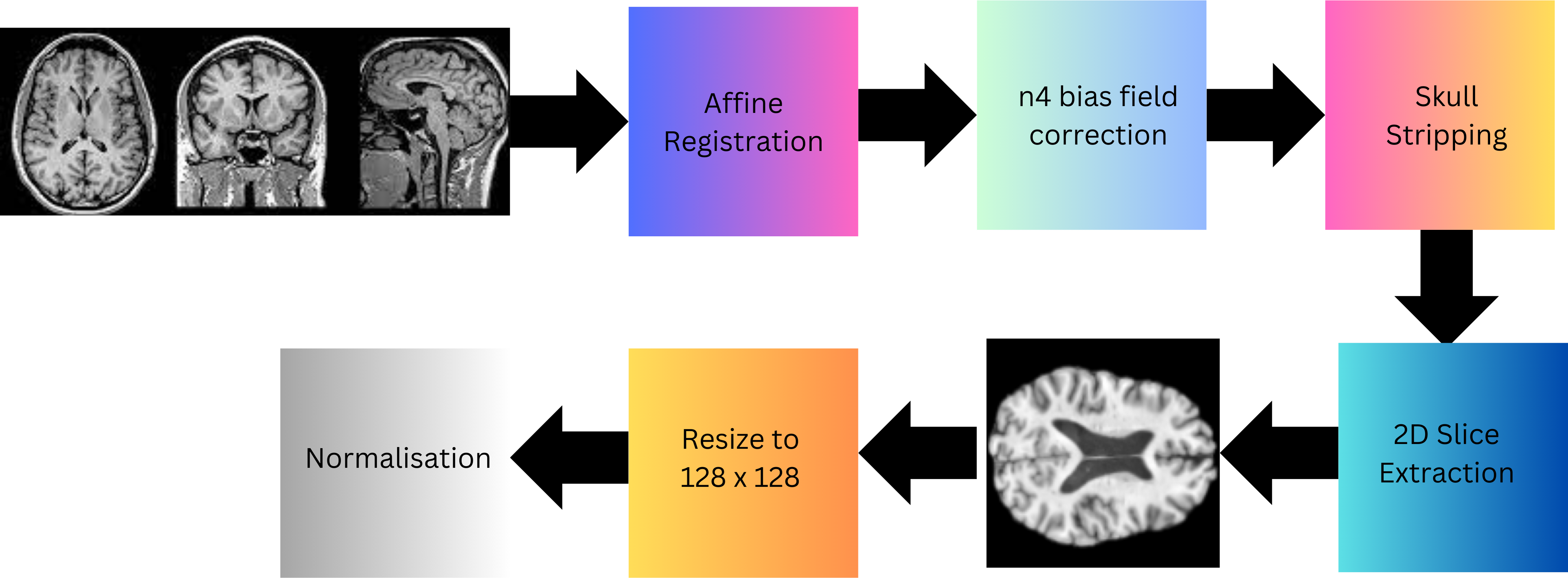}
    \caption{Our pre-processing pipeline. All pre-processing was conducted in the Python programming language using the HD bet and Simple ITK packages.}
    \label{prepro}
\end{figure}

Figure \ref{prepro} shows our pre-processing pipeline. All images started as 3D volumes. We first affinely registered the images to the MNI 152 brain template using the ants package. Following this they were resampled to 130 × 130 × 130 resolution, and n4 bias field corrected with the Simple ITK package. The images were then skull stripped using HD bet and 2D medial axial slices were extracted. The final steps were to resize the images to 128 × 128 and normalised their pixel values to be between 0 and 1.
\subsection{Experimental setting} 
For generative benchmarks, we compared our approach against a diffusion autoencoder (DiffAE),  cVAE, and a VAE. We computed both our latent similarity metric and image space deviation metrics. We standardised the distributions of both metrics to have a mean of zero and a standard deviation of one. As per Pinaya et al. (2021) \cite{walternature} our image space deviation metric was the MSE between images from our ALS cohort and their reconstructions. Where higher MSEs are indicative of being further away from the healthy cohort. We refer to latent similarity metrics and image space deviation metrics jointly as a \textit{normative scores} for brevity. For non-generative benchmarks, we compared our approach against machine learning models trained to predict age as a pretext task and then computed normative scores. All models were parameter matched and all analyses were conducted in the Python programming language. 
\subsubsection{Network Architecture, Hardware, and Model Training}
\begin{figure}[t]
    \centering 
    \includegraphics[scale=0.1]{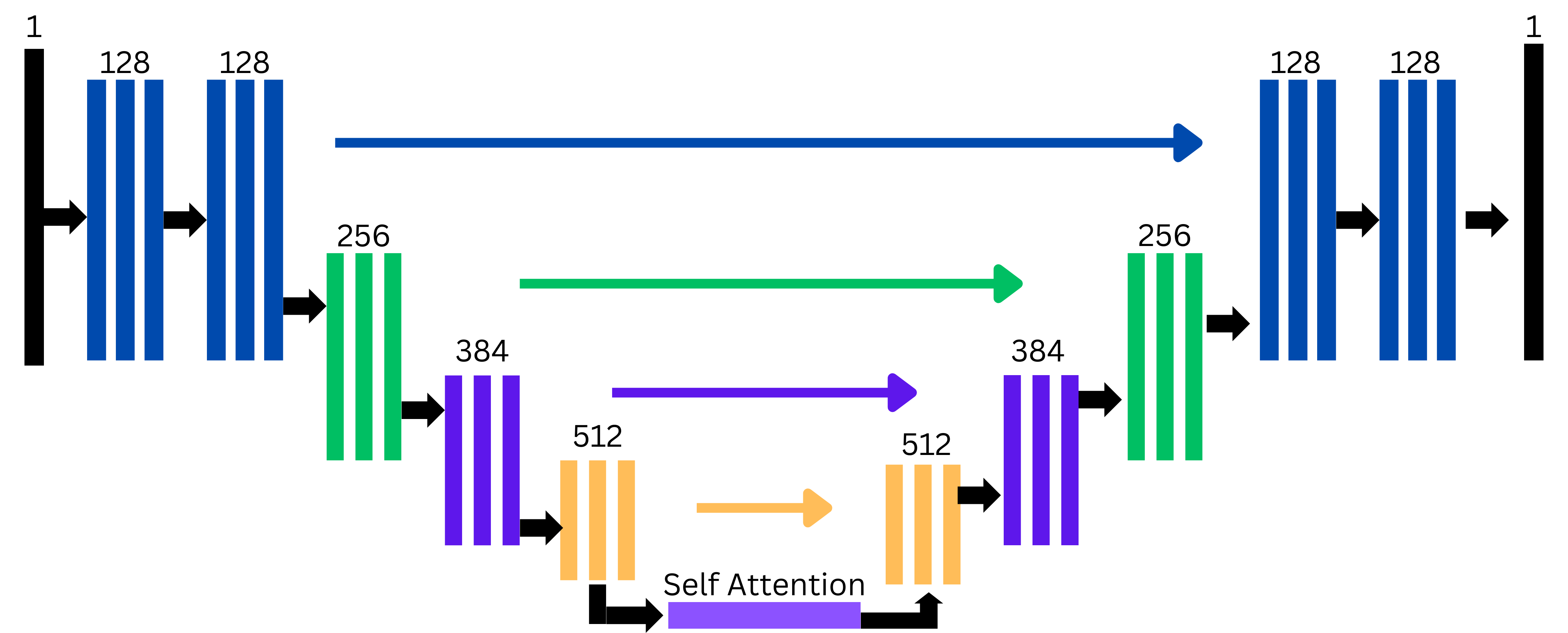}
    \caption{The architure of our noise predictor model, $\epsilon_{\theta}^{(t)}(\mathbf{x}_{t},\mathbf{z})$. Each set of bars are convolutional layers, and the numbers above denote the output channels.}
    \label{archi}
\end{figure}

Our noise predictor network $\epsilon_{\theta}^{(t)}(\mathbf{x}_{t},\mathbf{z})$ had the U-Net architecture seen in figure \ref{archi}, whilst $s_{\phi}(\mathbf{x}_{0}, \alpha)$ was only the downward path. Group normalisation and the SiLU activation function followed each neural network layer. There were 123 million parameters in total. The VAEs in our benchmarks followed the same channel expansions as $\epsilon_{\theta}^{(t)}(\mathbf{x}_{t},\mathbf{z})$ . All training was performed on an Nvidia GeForce RTX 4090 graphics card, with the Adam optimizer using PyTorch lightning. The lifelines package \cite{lifelines} was used for survival analysis and the scipy package was used for statistical analyses \cite{Virtanen2020SciPyPython}. 
\subsection{Results}
\subsubsection{Association with survival time.} 
\begin{figure}[t]
    \centering 
    \includegraphics[scale=0.25]{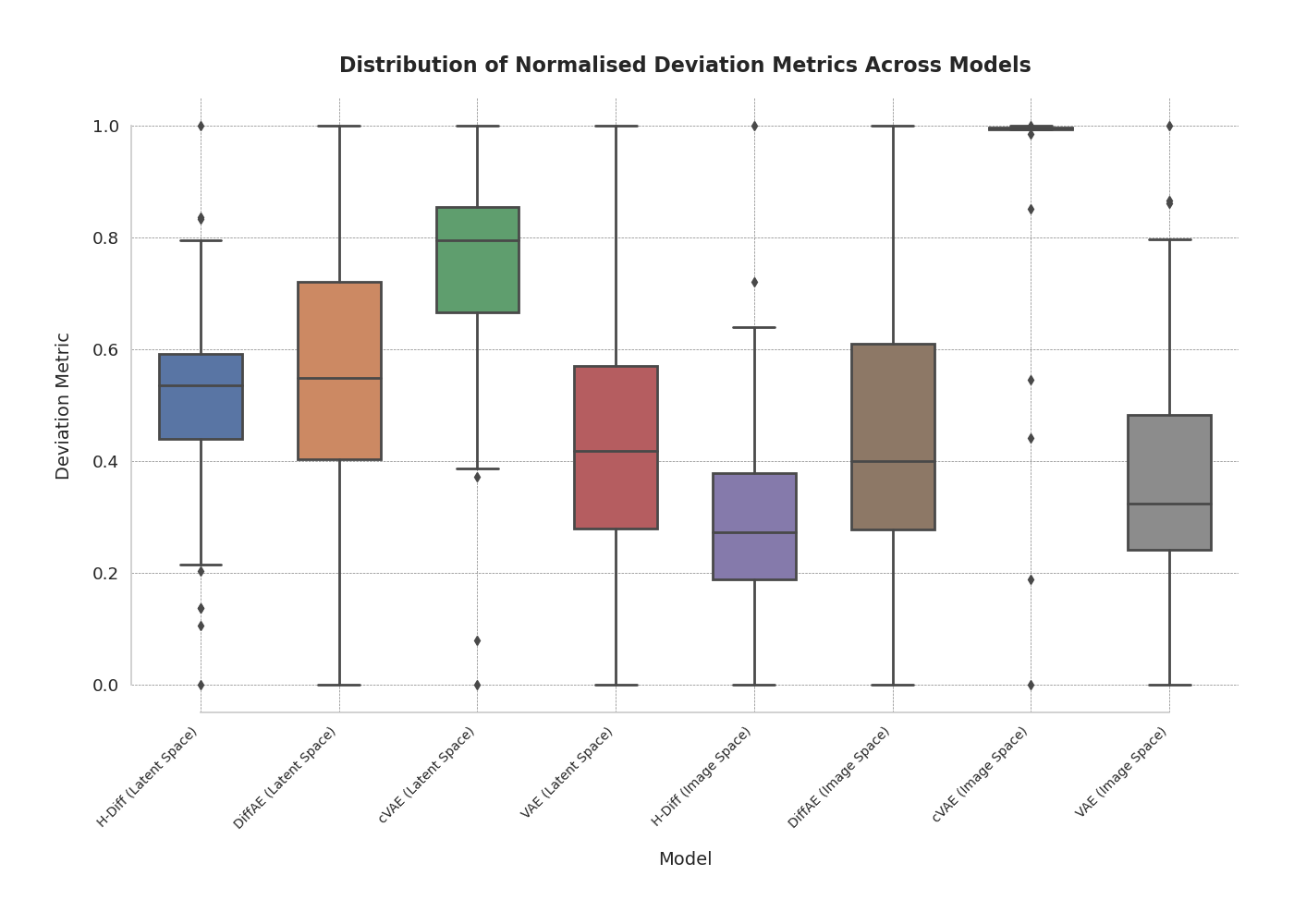}
    \caption{The distributions of normative scores across different models. There values were normalised to be between 0 and 1.}
    \label{boxxie}
\end{figure}

We assessed the linear association between our latent similarity metric and survival time in days. Pearson's r and Kendall's Tau both yielded statistically significant correlations of 0.34 p<0.05 and 0.28 p<0.001 respectively. Table \ref{table1} displays how this compares to a DiffAE, cVAE, and a VAE. The results show that our hierarchical diffusion autoencoder (H-Diff) has the strongest association with survival across those two metrics. We also report the normalised mutual information and the equivalent results for the image space deviation metric. Figure \ref{boxxie} shows the distribution of the normative scores across different models. 
\begin{table}
\centering
\caption{Association with survival time. The first three columns correspond to the association between our latent similarity metric and survival time across different models. The final three columns correspond to the association between the image space deviation metric and survival time. The normalised mutual information is abbreviated as NMI. Significant results (p<0.05) are in \textbf{bold}.}
\begin{tabular}{|l|l|l|l|l|l|l|}
\hline
& \multicolumn{3}{c|}{LATENT-SPACE} & \multicolumn{3}{c|}{IMAGE-SPACE} \\
\cline{2-7}
 & Pearson's R & Kendall's Tau & NMI & Pearson's R & Kendall's Tau & NMI \\ 
\hline
\textbf{H-Diff} & \textbf{0.34} & \textbf{0.28} & 0.25 & \textbf{0.24} & \textbf{0.20} & 0.20 \\
\hline 
DiffAE & -0.11   & 0.10  & 0.24 & \textbf{0.26}& \textbf{0.21} & 0.23 \\ 
\hline
cVAE & -0.16  & \textbf{-0.16} & 0.20 & 0.17  & \textbf{0.24} & 0.10 \\ 
\hline
VAE & \textbf{-0.31} & \textbf{-0.24} & 0.25 & \textbf{0.23} & \textbf{0.19} & 0.24 \\ 
\hline 
\end{tabular}
\label{table1} 
\end{table}
\subsubsection{Cox Proportional Hazard Models.} 
We fit Cox proportional hazards models to our latent similarity metric with age and sex as covariates. Our metric yielded a statistically significant hazard ratio of 0.73 p<0.05 with a 95\% CI of [0.57-0.94]. This means that a unit increase in cosine similarity is associated with a 27\% reduction in the risk of mortality occurring at any time point, assuming all other factors remain constant. Due to the standardisation procedure we followed, a unit increase is equivalent to an increase of one standard deviation away from the mean of our ALS latent similarity distribution. Table \ref{cox1} displays how this compares to other generative approaches across normative scores.

To compare our method to non-generative techniques we used pre-trained machine learning models which predict an individual's age based on their MR scan. We then calculated the difference between their predicted age and chronological age to produce brain-predicted age differences (brain-PAD). This deviation metric is associated with survival time in ALS and neurodegenerative disease in previous works \cite{ijishakin2023semi,Fran,DeepBrainNet}. Table \ref{cox2} displays the results of Cox models using age, sex, and the brain PADs on our ALS cohort from various approaches. Brain-PADs were not significant predictors of survival in contrast to the present method.

\begin{figure*}[t]
\centering
    \includegraphics[scale=0.111]{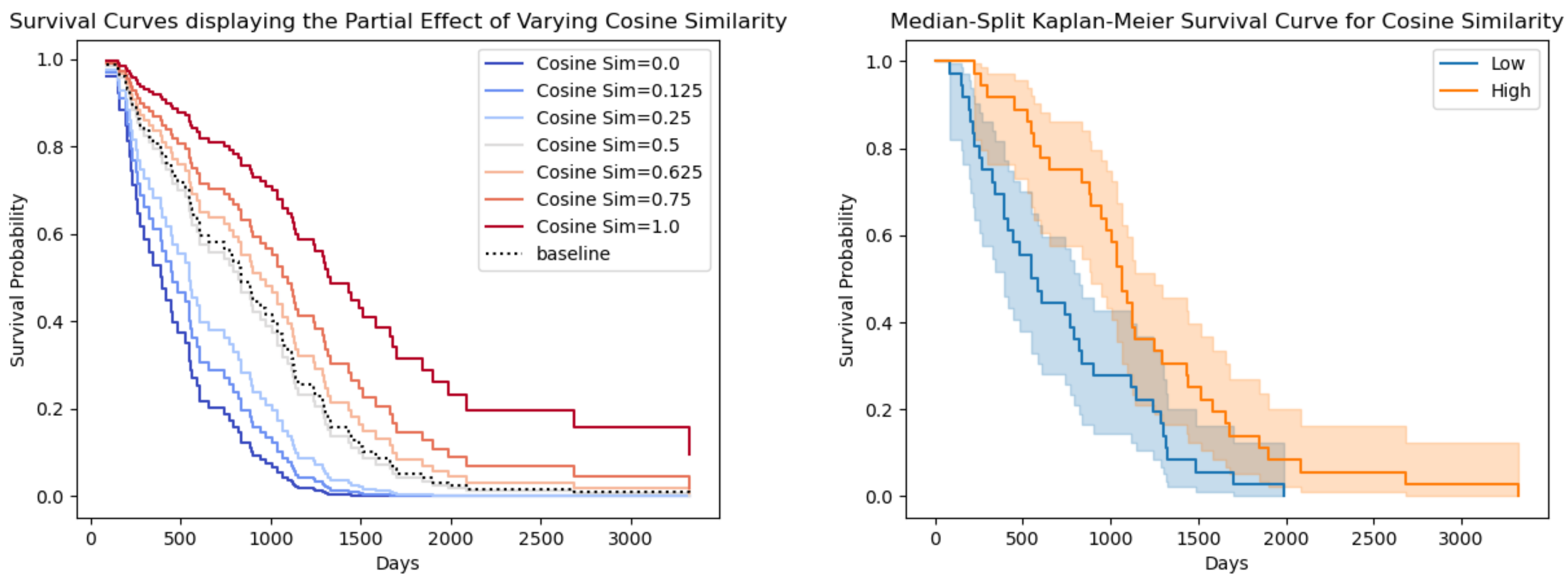}
\caption{Survival curves (Kaplan-Meier estimators) as a function of varying cosine similarity. On the left is a partial effect model, varying cosine similarity, and holding all else equal. On the right are the survival curves across a median split of cosine similarity.} 
\label{survival}
\end{figure*} 
\begin{table}[]
\centering
\caption{Results of Cox models using both image space and latent space based normative scores. Each row denotes a Cox model where the normative score was derived from a different generative model. All results are reported as Hazard ratio (HR) [95\% CI] and significant results (p < 0.05) are in \textbf{bold}.}
\begin{tabular}{|l|l|l|l|l|l|}
\hline
Model & Sex HR [CI] & Age HR [CI] &  Normative Score HR [CI] & Score Type \\ 
\hline 
H-Diff & 1.17 [0.70-1.95] & 1.00 [0.99-1.03] & \textbf{0.73 [0.57-0.94]} & Latent-Space \\
H-Diff & 1.21 [0.71-2.06] & 1.00 [0.98-1.03] & 0.83 [0.61-1.12] & Image-Space \\
\hline 
DiffAE & 1.38 [0.86-2.23] & 1.01 [0.99-1.03] & 1.14 [0.91-1.43] & Latent-Space \\
DiffAE & 1.22 [0.68-1.98] & 1.00 [0.98-1.03] & 0.80 [0.60-1.06] & Image-Space \\
\hline 
cVAE & 1.34 [0.81-2.26] & 1.01 [0.98-1.03] & 1.08 [0.83-1.41] & Latent-Space \\
cVAE & 1.41 [0.87-2.28] & 1.01 [0.98-1.03] & \textbf{0.77 [0.61-0.99]} & Image-Space \\
\hline 
VAE & 1.15 [0.70-1.90] & 1.00 [0.98-1.03] & \textbf{1.40 [1.07-1.84]} & Latent-Space \\
VAE & 1.25 [0.74-2.12] & 1.00 [0.99-1.03] & 0.87 [0.68-1.13] & Image-Space \\
\hline
\end{tabular}
\label{cox1}
\end{table}
\begin{table}[]
\centering
\caption{Results of Cox models with non-generative approaches. The rows denote the machine learning model used to produce the brain PADs featured in the hazard regression. All results are reported as Hazard Ratio [CI]. None of the models contained significant predictors.}
\begin{tabular}{|l|l|l|l|l|}
\hline
Model           & Sex HR [CI]   & Age HR [CI]   & Brain-PAD HR [CI] \\ 
\hline
SS-Diffusion \cite{ijishakin2023semi}   & 0.69 [0.42-1.12] & 1.00 [0.98-1.03] & 0.99 [0.98-1.01]\\
\hline
MIDI BrainAge \cite{Wood2022AccurateExaminations}   & 0.73 [0.45-1.17] & 1.02 [0.99-1.05] & 1.02 [0.98-1.06] \\
\hline
brainageR \cite{Fran}     & 0.72 [0.44-1.18]  & 1.00 [0.96-1.05] & 0.99 [0.95-1.03] \\
\hline
DeepBrainNet \cite{DeepBrainNet}   & 0.69 [0.43-1.12]  & 1.01 [0.99-1.04] & 1.02 [0.99-1.05]  \\
\hline
\end{tabular}
\label{cox2} 
\end{table} 
\subsubsection{Survival Plots and Kolmogorov Smirnov Tests.}
Figure \ref{survival} shows survival curves (modeled with Kaplan Meier estimators) as a function of varying cosine similarity. 
% One curve displays in a partial effects model showing what the overall survival curve would look like if the cosine similarity were increased or decreased holding all else equal. Whilst the other is a Kaplain-meier plot, showing the survival curves across a median split of cosine similarity. 
Both, curves demonstrate that higher cosine similarity is strongly associated with higher survival times. A two-sample Kolmogorov-Smirnov (KS) test, applied after dividing survival times by the median of the cosine similarity levels, revealed a significant discrepancy between the groups, with a KS test statistic of 0.44 (p < 0.05). Additionally, when the cosine similarity levels themselves were split at their median and another two-sample KS test was applied it yielded a KS test statistic of 0.90 (p < 0.001). These results demonstrate that the two groups differ significantly.

\section{Discussion and Future Work} 
We presented the normative diffusion autoencoder framework, alongside a new class of diffusion model called the hierarchical diffusion autoencoder. Our model surpassed both its generative and non-generative counterparts in producing a strong and interpretable predictor of survival time in ALS. With a unit increase in our metric being associated with a 27\% reduction in risk of mortality. In other words, if a patient's brain structure in the latent space was within 1 standard deviation of the healthy cohort mean, mortality risk over the study period (9.6 years) is reduced by over a quarter. Although other models produced strong predictors of survival they were either predictive in an uninterpretable direction in that images closer to the healthy cohort had shorter survival lengths or were not as strongly associated. As such, our method represents a promising move toward producing interpretable and powerful deep normative modelling-derived biomarkers for ALS prognostication. Future work should aim to extend this approach to fully 3D MRI data, and also remove the explicit separation between the training and survival analysis phases.  

% \printbibliography 
\bibliographystyle{splncs03_unsrt} 
% \bibliography{mybib} 

\begin{thebibliography}{28}
\bibitem{globalprev}
Wolfson, C., Gauvin, D., Ishola, F., Oskoui, M.: The global prevalence and incidence of amyotrophic lateral sclerosis: A systematic review (p2-5.023). Neurology 100 (17 Supplement 2) (2023)

\bibitem{Knibb2016AALS}
Knibb, J.A., Keren, N., Kulka, A., et al.: A clinical tool for predicting survival in ALS. Journal of Neurology, Neurosurgery and Psychiatry 87(12), 1361--1367 (2016). doi:10.1136/JNNP-2015-312908

\bibitem{ijishakin2023semi}
Ijishakin, A., Martin, S.A., Townend, F.J., Cole, J.H., Malaspina, A.: Semi-supervised diffusion model for brain age prediction. In: Deep Generative Models for Health Workshop NeurIPS 2023 (2023)

\bibitem{Swinnen2014TheSclerosis}
Swinnen, B., Robberecht, W.: The phenotypic variability of amyotrophic lateral sclerosis. Nature Reviews Neurology 10, 661--670 (2014). doi:10.1038/nrneurol.2014.184. Available: www.nature.com/nrneurol

\bibitem{agosta2010present}
Agosta, F., Chiò, A., Cosottini, M., et al.: The present and the future of neuroimaging in amyotrophic lateral sclerosis. American Journal of Neuroradiology 31(10), 1769--1777 (2010)

\bibitem{bozek2023normative}
Bozek, J., Griffanti, L., Lau, S., Jenkinson, M.: Normative models for neuroimaging markers: Impact of model selection, sample size and evaluation criteria. NeuroImage 268, 119864 (2023)

\bibitem{wolfers2018mapping}
Wolfers, T., Doan, N.T., Kaufmann, T., et al.: Mapping the heterogeneous phenotype of schizophrenia and bipolar disorder using normative models. JAMA Psychiatry 75 (11), 1146--1155 (2018)

\bibitem{zabihi2019dissecting}
Zabihi, M., Oldehinkel, M., Wolfers, T., et al.: Dissecting the heterogeneous cortical anatomy of autism spectrum disorder using normative models. Biological Psychiatry: Cognitive Neuroscience and Neuroimaging 4 (6), 567--578 (2019)

\bibitem{rutherford2022charting}
Rutherford, S., Fraza, C., Dinga, R., et al.: Charting brain growth and aging at high spatial precision. eLife 11, e72904 (2022)

\bibitem{bethlehem2022brain}
Bethlehem, R.A., Seidlitz, J., White, S.R., et al.: Brain charts for the human lifespan. Nature 604 (7906), 525--533 (2022)

\bibitem{borghi2006construction}
Borghi, E., de Onis, M., Garza, C., et al.: Construction of the world health organization child growth standards: Selection of methods for attained growth curves. Statistics in Medicine 25 (2), 247--265 (2006)

\bibitem{marquand2019conceptualizing}
Marquand, A.F., Kia, S.M., Zabihi, M., Wolfers, T., Buitelaar, J.K., Beckmann, C.F.: Conceptualizing mental disorders as deviations from normative functioning. Molecular Psychiatry 24 (10), 1415--1424 (2019)

\bibitem{yee2015vector}
Yee, T.W.: Vector Generalized Linear and Additive Models: With an Implementation in R. Springer (2015)

\bibitem{Ho}
Ho, J., Jain, A., Abbeel, P.: Denoising diffusion probabilistic models. Advances in Neural Information Processing Systems 33, 6840--6851 (2020)

\bibitem{song}
Song, J., Meng, C., Ermon, S.: Denoising diffusion implicit models. arXiv preprint arXiv:2010.02502 (2020)

\bibitem{DiffAE}
Preechakul, K., Chatthee, N., Wizadwongsa, S., Suwajanakorn, S.: Diffusion autoencoders: Toward a meaningful and decodable representation. In: 2022 IEEE/CVF Conference on Computer Vision and Pattern Recognition (CVPR), pp. 10609--10619. IEEE (2021)

\bibitem{wang2023}
Wang, X., Zhou, R., Zhao, K., Leow, A., Zhang, Y., He, L.: Normative modeling via conditional variational autoencoder and adversarial learning to identify brain dysfunction in Alzheimer’s disease. In: 2023 IEEE 20th International Symposium on Biomedical Imaging (ISBI), pp. 1--4. IEEE (2023)

\bibitem{ana2022}
Lawry Aguila, A., Chapman, J., Janahi, M., Altmann, A.: Conditional VAEs for confound removal and normative modelling of neurodegenerative diseases. In: International Conference on Medical Image Computing and Computer-Assisted Intervention, pp. 430--440. Springer (2022)

\bibitem{ana2023}
Aguila, A.L., Chapman, J., Altmann, A.: Multi-modal variational autoencoders for normative modelling across multiple imaging modalities. arXiv preprint arXiv:2303.12706 (2023)

\bibitem{Walter}
Pinaya, W.H., Graham, M.S., Gray, R., et al.: Fast unsupervised brain anomaly detection and segmentation with diffusion models. In: International Conference on Medical Image Computing and Computer-Assisted Intervention, pp. 705--714. Springer (2022)

\bibitem{patchy-diffusion}
Behrendt, F., Bhattacharya, D., Krüger, J., Opfer, R., Schlaefer, A.: Patched diffusion models for unsupervised anomaly detection in brain MRI. In: Medical Imaging with Deep Learning, pp. 1019--1032. PMLR (2024)

\bibitem{iqbal}
Iqbal, H., Khalid, U., Chen, C., Hua, J.: Unsupervised anomaly detection in medical images using masked diffusion model. In: International Workshop on Machine Learning in Medical Imaging, pp. 372--381. Springer (2023)

\bibitem{walternature}
Pinaya, W.H., Scarpazza, C., Garcia-Dias, R., et al.: Using normative modelling to detect disease progression in mild cognitive impairment and Alzheimer’s disease in a cross-sectional multi-cohort study. Scientific Reports 11 (1), 15746 (2021)

\bibitem{lifelines}
Davidson-Pilon, C.: Lifelines: Survival analysis in Python. Journal of Open Source Software 4 (40), 1317 (2019)

\bibitem{Virtanen2020SciPyPython}
Virtanen, P., Gommers, R., Oliphant, T.E., et al.: SciPy 1.0: Fundamental algorithms for scientific computing in Python. Nature Methods 17 (3), 261--272 (2020)

\bibitem{Fran}
Biondo, F., Jewell, A., Pritchard, M., et al.: Brain-age is associated with progression to dementia in memory clinic patients. NeuroImage: Clinical 36, 103175 (2022)

\bibitem{DeepBrainNet}
Bashyam, V.M., Erus, G., Doshi, J., et al.: MRI signatures of brain age and disease over the lifespan based on a deep brain network and 14468 individuals worldwide. Brain 143 (7), 2312--2324 (2020)

\bibitem{Wood2022AccurateExaminations}
Wood, D.A., Kafiabadi, S., Busaidi, A.A., et al.: Accurate brain-age models for routine clinical MRI examinations. NeuroImage 249, 118871 (2022). doi:10.1016/J.NEUROIMAGE.2022.118871

\end{thebibliography}

\end{document}

\title{Supplementary Material}

\author{Anonymous}

\institute{Anonymous Organization \\\email{**@******.***}}

\maketitle 
\begin{figure}
    \centering 
    \includegraphics[scale=0.06]{images/ALS Dataset.png}
    \caption{Our dataset. On the left is a description of our ALS cohort, and on the right are the names of the 7 datasets that comprised our healthy cohort.}
    \label{}
\end{figure} 

\begin{figure}
    \centering 
    \includegraphics[scale=0.06]{images/Affine Transformation (1).png}
    \caption{Our pre-processing pipeline. All images started as 3D volumes. We first affinely registered the images to the MNI 152 brain template using the ants package. Following this they were resampled to 130 × 130 × 130 resolution, and n4 bias field corrected with the Simple ITK package. The images were then skull stripped using HD bet and 2D medial axial slices were extracted. The final steps were to resize the images to 128 × 128 and normalised their pixel values to be between 0 and 1.}
    \label{}
\end{figure} 
\clearpage 
\begin{figure}[]
    \centering 
    \includegraphics[scale=0.1]{images/1 (3).png}
    \caption{Model architecture. Our noise predictor network $\epsilon_{\theta}^{(t)}(\mathbf{x}_{t},\mathbf{z})$ had the architecture in the figure, whilst $s_{\phi}(\mathbf{x}_{0}, \alpha)$ was only the downward path. Each set of bars are convolutional layers, and the numbers above denote the output channels. Each layer was followed by group normalisation and the SiLU activation function. There were 123 million parameters in total. The VAEs in our benchmarks followed the same channel expansions displayed above and were parameter matched.  All training was performed on an Nvidia GeForce RTX 4090 graphics card, with the Adam optimizer using PyTorch lightning.}
    \label{}
\end{figure}